\documentclass[10pt, conference, compsocconf]{IEEEtran}
\usepackage{graphicx}
\usepackage{tikz}
\usepackage{tikzpagenodes}
\usetikzlibrary{shapes,arrows,shapes.geometric,fit}
\usepackage{pgfpages}
\usepackage{subcaption}
\usepackage{pgfplots}

\usepackage{algorithm}
\usepackage{algorithmicx}
\usepackage[noend]{algpseudocode}
\usepackage{mathtools}
\usepackage{wasysym}
\usepackage{stmaryrd}

\usepackage{url}
\usepackage[bookmarks=false]{hyperref}
\usepackage{amsfonts}
\usepackage{amssymb}
\usepackage{etoolbox}
\usepackage{fancyhdr}
\PassOptionsToPackage{table}{xcolor}
\usepackage{xcolor}
\usepackage{colortbl}
\usepackage{multirow}
\usepackage{comment}
\usepackage{framed}
\usepackage{epigraph}
\usepackage{balance}  
\usepackage{amssymb}
\usepackage{soul}
\usepackage{pgfplots}
\usetikzlibrary{plotmarks}

\pgfplotsset{
  compat=newest,
  xlabel near ticks,
  ylabel near ticks
}

\newcommand\openstar[1][0.6]{%
  \scalerel*{%
    \stackinset{c}{-.125pt}{c}{}{\scalebox{#1}{\color{white}{$\star$}}}{%
      $\star$}%
  }{\star}
}

\def\angle{20}
\def\radius{3}
\def\cyclelist{{"gray","white"}}
\newcount\cyclecount \cyclecount=-1
\newcount\ind \ind=-1

\pgfmathdeclarefunction{gauss}{2}{%
  \pgfmathparse{1/(#2*sqrt(2*pi))*exp(-((x-#1)^2)/(2*#2^2))}%
}

\usepackage{subcaption}
\DeclareCaptionLabelSeparator{periodspace}{.\quad}
\begin{document}

\title{Show me the material evidence --- Initial experiments\\ on evaluating hypotheses from user-generated multimedia data}

\author{
\\[-31pt]
\IEEEauthorblockN{Bernardo Gon\c{c}alves}
\IEEEauthorblockA{IBM Research\\
S\~ao Paulo, Brazil\\
begoncalves@acm.org}
\\[-22pt]
}

\maketitle

\begin{abstract}
Subjective questions such as `does neymar dive', or `is clinton lying', or `is trump a fascist', are popular queries to web search engines, as can be seen by autocompletion suggestions on Google, Yahoo and Bing. In the era of cognitive computing, beyond search, they could be handled as hypotheses issued for evaluation. Our vision is to leverage on unstructured data and metadata of the rich user-generated multimedia that is often shared as material evidence in favor or against hypotheses in social media platforms. In this paper we present two preliminary experiments along those lines and discuss challenges for a cognitive computing system that collects material evidence from user-generated multimedia towards aggregating it into some form of collective decision on the hypothesis. 
\end{abstract}


%
\IEEEpeerreviewmaketitle

  \begin{tikzpicture}[remember picture,overlay]
    \node[align=center] at ([yshift=-2.0em]current page text area.south) {\footnotesize In: \emph{Proc. of the 1st Workshop on Multimedia Support for Decision-Making Processes, at IEEE Intl. Symposium on Multimedia (ISM'16)}, San Jose, CA, 2016.};
  \end{tikzpicture}%

\section{Introduction}\label{sec:intro}

\noindent
Recent advances in artificial intelligence and other areas are enabling a generation of systems so-called `cognitive',\footnote{See [\url{http://en.wikipedia.org/wiki/Cognitive_computing}].} for their capability to digest large volumes of unstructured data and come up with an answer to some interesting question \cite{ferrucci2010}, such as whether one's favorite soccer player `dives' or does not `dive' according to the general public opinion.\footnote{Soccer players are said to `dive' if they frequently playact unjustified falls onto the pitch.}  

In fact, such opinionated claims, whether in affirmative, negative or interrogative form, are frequent in news feeds, microblogs and social media platforms. They are produced by an army of journalists, blog writers and ordinary people through their social media accounts, all in making sense of the world with varying degrees of analytical depth.

For many such hypotheses or decision questions, however, some kind of multimedia content is expected to both tie up and substantiate people's opinions, otherwise too abstract or uncompelling to be relevant to others. That user-generated content is meant to support or explain an opinion. Interestingly, people really seem to be willing to share such content as a form of `material evidence' in favor or against a hypothesis. 
For instance, consider the question `does Neymar dive?'. Fig. 1 shows top-ranked video content returned by Google from its social media platform as search results for the hypothesis about Neymar. Specifically, in 2013 a blog writer took the effort to manually collect and rate some material evidence (user-generated videos) on the hypothesis about Neymar, like in a citizen journalism investigation.\footnote{See article at [\url{http://www.givemesport.com/355414-does-neymar-dive}]. Retrieved on September 24, 2016.} 

The key insight behind this kind of procedure towards the evaluation of subjective hypotheses is that, although people may engage into never-ending discussions on such subjective topics, maybe on the interpretation of a particular piece of material evidence they can find themselves in agreement. At scale, if a data system accumulates material evidence provided by several users, preferably of balanced prior biases to meet a condition for the `wisdom of crowds' \cite{surowiecki2004}, then cognitive computing could enable analytics and a decision for answering such interesting subjective questions. 

In this paper we pursue a ``baby'' step on the exploration of this idea by means of two preliminary experiments. They will both concentrate on the example hypothesis about Neymar. In one experiment, we have employed off-the-shelf web mining and text analytics to collect candidate claims on a named entity and detect its polarity, e.g., `Neymar' is (not) a `diver.' Although this kind of experiment has already been tried for fairly less controversial topics (like `cute animals,' `safe cities,' etc, see \cite{trummer2015}), we are not aware of any work on claim extraction and parsing that also observes whether the claim is attached to some (user-generated) multimedia as a form of material evidence to back it up. We then compare the proportion of positive/negative (supportive/skeptical) claims on the hypothesis, first with and without material evidence, and second against similar entities --- specifically, claims on Neymar are compared to claims on Lionel Messi and Cristiano Ronaldo. We report on the differences we have found, which for a caveat to the reader, are only worth to suggest further research. 

In a second and independent experiment, a sample of videos on the same hypothesis (Neymar's dives) has been collected and submitted to crowd workers for rating whether Neymar is diving or not. Before rating any video, the workers have been required to declare their prior biases towards the hypothesis. We then compare their video ratings with their declared prior biases, and find some shift towards centrality. Of course, questions may be raised about sample selection --- on which we have strived for balance, having included social media posts from apparently both polarities --- but again for a caveat, even though we have thought through our experiment design, the results are preliminary only, to ``bootstrap'' research. 

After presenting the two initial experiments in \S\ref{sec:web-claims}--\ref{sec:crowd-claims}, we proceed to a discussion of research challenges in \S\ref{sec:challenges}. Given the evidence that has been accumulated in its multiple forms (e.g., text and video), an interesting and core research problem is how to combine them into a weighted sum of scores towards an up-to-date hypothesis decision. 
In \S\ref{sec:related-work} we comment on related work, and in \S\ref{sec:conclusions} we conclude the paper. 

\begin{figure}[t]
\includegraphics[width=.43\textwidth]{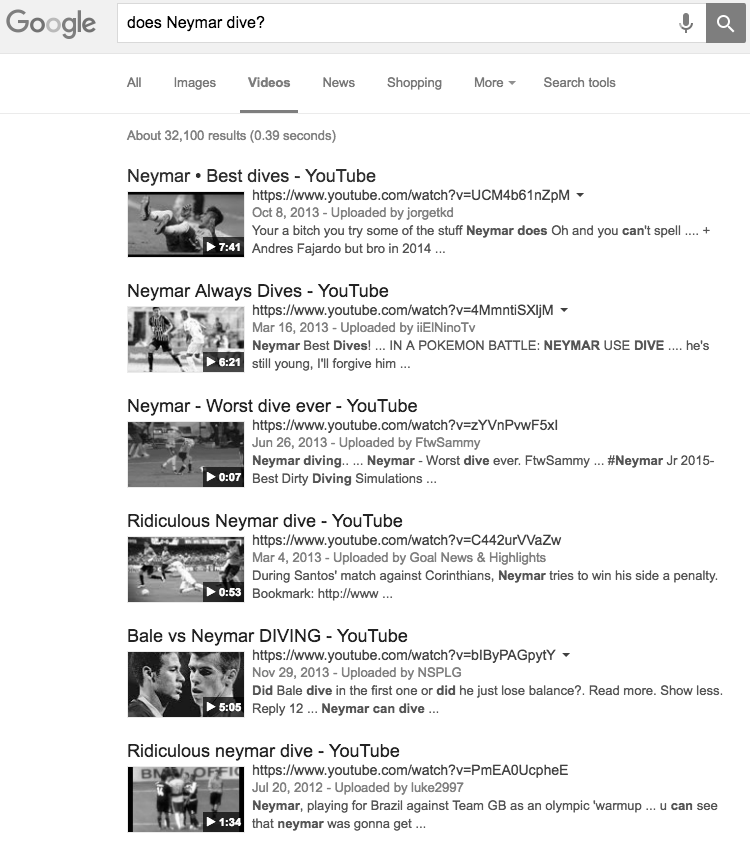}
\caption{Video search results on the hypothesis about Neymar.}
\label{fig:google}
\vspace{-10pt}
\end{figure}

\section{Experiment 1: Parsing Claims from the Web}\label{sec:web-claims}
\noindent
We start the section by introducing the state-of-the-art cognitive technologies that we have used as well as our data collection process. Then we present the initial results.

\subsection{Data Collection from the Web}
\noindent
We will rely on social media data retrieved from a popular search engine. Specifically, we have sent the hypothesis question about Neymar for search in 3 different forms, affirmative, negative and interrogative inflections, (resp.) `\emph{Neymar is a diver},' `\emph{Neymar is not a diver},' and `\emph{does Neymar dive?}' Among the 3 sets of search results obtained, some results will overlap. We combine the result sets by union, eliminating duplicates. Moreover, we will repeat the search process 2 times by restricting it to each of two modalities, text and video. The result sets are then combined by union as well, with the media type kept in record).

Once we have in our database such social media data links/entries that are related to the target question (deemed relevant by the search engine), we carry out some cognitive processing on them --- essentially, a Natural Language Processing (NLP) method for claim parsing that we describe shortly. Those data entries will always be uniquely identified by their \emph{url}, and be composed of a \emph{title} and a \emph{snippet} that will be object of the NLP, starting with the title. 

From each such data entry, if a claim is validated then it is considered a form of evidence (one count, either positive or negative) in favor or against the hypothesis. If the entry is of video modality, in particular, then it is considered to have been offered as `material' evidence for the counted claim on the hypothesis. We shall then compare the aggregated results whose sources are of the different modalities, text and video. 

\subsection{NLP-based Parsing and Validation of Claims}\label{subsec:parsing}
\noindent
We use some of the state-of-the-art web mining and text analytics technologies that are available, for the extraction of candidate claims from the web and their validation as referring to the target hypothesis. 

\begin{figure}\centering
\includegraphics[width=.35\textwidth]{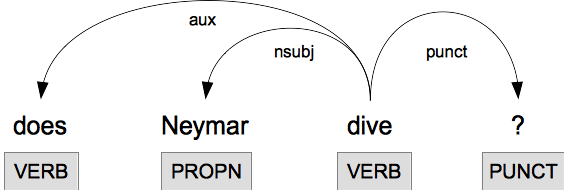}
\caption{Parsed question about Neymar. 
Terms are shown with their part-of-speech tag, and connected in a linguistic dependency parse tree labeled according to the CLEAR style \cite{choi2012}. Term token `dive' is the root element of the dependency parse tree, having term token `Neymar' as a child.}
\label{fig:neymar-query}
\end{figure}

In particular, for the web mining task we have used \textsf{Pattern}.\footnote{\textsf{Pattern}, available at [\url{http://www.clips.ua.ac.be/pattern}].} 
For the NLP core tasks we have used \textsf{spaCy},\footnote{\textsf{spaCy}, available at [\url{http://spacy.io/}]. Retrieved on Sep. 24, 2016.} which is a fast and accurate API for several NLP tasks \cite{honnibal2015,choi2015}, specially dependency parsing which is key for us. 
For our example hypothesis, ``\emph{does Neymar dive?},'' the dependency parse tree rendered by \textsf{spaCy} is shown in Fig.~\ref{fig:neymar-query}. For each data link/entry that has a valid claim, the output that we store is illustrated in Table.~\ref{tab:claim}. 

\begin{table*}
\caption{Example of claims that we have detected and analyzed by means of linguistic dependency parse trees. The positive polarity value is detected by parsing the term token dependencies from the sentence. }
\label{tab:claim}
\vspace{-5pt}
\begin{center}
\begingroup\setlength{\fboxsep}{1pt}
\colorbox{gray!8}{%
\begin{tabular}{c | c | p{0.55\textwidth} | c }
$\!\!$URL$\!$ & MEDIA & CLAIM & POSITIVE\\
 \hline
\href{http://11x2.com/news/1696074/match-highlights-la-liga-bbva-week-28-barcelona-2-1-real-madrid-22-3-15}{[http://tinyurl.com/hn4hndq]} & text & $\!$22 Mar 2015 ... When did Neymar dive in that match ? .... Neymar right after Mathies goal, then Ronaldo hitting the crossbar, and both Bravo and Casillas ...$\!\!$ & false\\
 \hline
\href{http://www.youtube.com/watch?v=5ePSZEg2AuY}{[http://tinyurl.com/jns5bt8]} & video & $\!$25 Sep 2013 - 3 sec - Uploaded by King Kong Neymar dive against Uruguay 2606 Neymar Dive Brazil vs Uruguay Confedcup 2013 Spain ...$\!\!$ & true\\
\end{tabular}
}\endgroup
\end{center}
\vspace{-15pt}
\end{table*}

\begin{figure}[t]
\includegraphics[width=.45\textwidth]{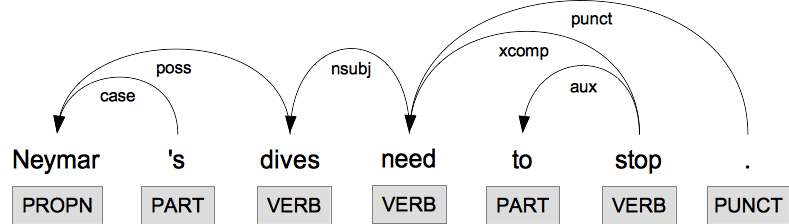}
\caption{Parsed sentence ``\emph{Neymar's dives need to stop.}''. Term token `Neymar' (target entity) is a child of term token `dives' (target property).}
\label{fig:neymar-phrase}
\vspace{-10pt}
\end{figure}

Given the natural language phrases that are extracted from the web, in order to disambiguate them onto the target conceptual structure `entity-has/(hasn't)-property,' our approach is as follows.
We try to locate both the entity and the property in a single sentence within a phrase, keeping a pointer to their tokens. If the entity and property do not appear together in a same sentence, a claim will not be identified. That is, in our current approach we look for sentence-level claims only and do not address coreference NLP tasks. Furthermore, even if they appear together, we check if their syntactic connection satisfy one of a few patterns that we expect.  
The analysis that is carried out at the sentence level goes on as follows. 

\textbf{Entity detection}. A term token is detected to stand for the target named entity in a sentence if all these conditions hold: (i) string similarity based on the Levenshtein edit distance function is higher than a threshold \cite{navarro2001}; and (ii) it is tagged as a proper name (PROPN) by part-of-speech processing.

\textbf{Property detection}. A term token is detected to stand for the target property if (i) string similarity on its lemma is higher than a threshold; \textsf{spaCy}'s lemmatization is based on Princeton's Wordnet. 
String similarity against the target property is only done once the candidate term token is already in its canonical lexical form. 

\begin{figure}[t]\centering
\includegraphics[width=.39\textwidth]{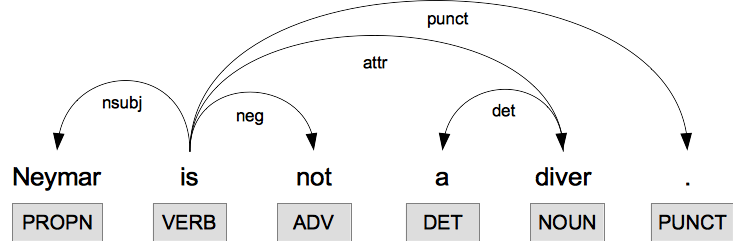}
\caption{Parsed sentence ``\emph{Neymar is not a diver.}''. Term tokens `Neymar' (target entity) and `diver' (target property) are siblings under `is.'}
\label{fig:neymar-phrase-alt}
\vspace{-8pt}
\end{figure}

\textbf{Entity-property structure detection}. The conceptual structure is validated if any one of these conditions hold: (i) the entity token is a child of the property token (e.g., the nominal subject (nsubj) of it, see Fig.~\ref{fig:neymar-query}; or has a possessive modifier (poss) dependency, see Fig.~\ref{fig:neymar-phrase}); or the entity and property tokens are siblings under verb to-be (see Fig.~\ref{fig:neymar-phrase-alt}).

Note that properties of interest (diver, lier, etc) can appear in real claims in either of a few forms, as:
\begin{itemize}
\item Verbs (e.g., ``\emph{Neymar \underline{dives} too much}''; note that `dive,' `diving,' `dived' all reduce to lemma `dive');
\item Nouns (e.g., ``\emph{Neymar's \underline{dive} last game},'' or ``\emph{Neymar is a \underline{diver}}.''.
\end{itemize}
\noindent
`Diver' (noun) is already in its lemma form, which is different from lemma `dive' (verb) yet they should be considered the same property, say, with canonical form `diver'.

\textbf{Negation detection}. Once a claim has been found/validated against an expected entity-property structure, we still need to detect its polarity. Although we look for claims out of interrogative sentences too, when processing results we only consider positive and negative claims, abstracting then the interrogative form (e.g., `\emph{does Neymar dive?}') as skepticism (a negative count) for the assertion of the property upon the entity. This is an important assumption in terms of statistical modeling, and it will be revisited later in \S\ref{sec:challenges}. 

For the polarity (binary) classification, we count how many ``negation'' occurrences are found from the named entity token node upwards in the parse tree, where ``negation'' can be any of the following linguistic patterns:

\begin{enumerate}
\item A term token in the tree path upwards is identified as a negation (neg) in a dependency relation;
\item A question mark character is identified as a term token;
\item A `why' term is identified as a token, imposing an affirmative effect on a question phrase that has a question mark, e.g., ``\emph{Why does Neymar dive?}.''
\end{enumerate}
\noindent
If the total counts give an even number then the polarity is positive. It is negative otherwise. 

While such heuristics are definitely not perfect, it seems to work well on most cases we could test and think of and then serve the purpose of collecting some initial data.


\subsection{Web-Mined Claims: Preliminary results}
\noindent
Table~\ref{tab:counts} shows the preliminary results of this experiment.  In addition to named entity Neymar, we have also run our pipeline for two similar named entities, Lionel Messi and Cristiano Ronaldo. Including these two other popular soccer players in the study provides us with a baseline.

Some observations about the data shown in Table~\ref{tab:counts} are:

\begin{itemize}
\item The number of validated claims is fairly lower than the number of url's returned by the search engine. This may be either because of too much precision over recall in our parsing of claims, or because too many search engine's results are not so related to the target claims. 

\item Comparing the positive polarity proportions of claims on text and video, the difference is (Neymar) +21\%, (Messi) +8\%, and (Ronaldo) -4\%. Interestingly, Neymar has the higher difference. It may be suggestive of, say, social science research on whether the `diver' claims on him involve more passion than on others. 
\item Messi's rates are remarkably low w.r.t. the two other players. This seems right, as he is known in the public domain for his ``resilience to keep up on his feet.'' 
\item Considering the arithmetic mean proportion of text and video claims, the player who has the higher `diver' rates is Ronaldo (94\%; against for 87.5\% Neymar). If Neymar is more famously associated with the `diver' label, this might be considered some kind of counterevidence. 
\end{itemize}

\begin{table}[t]
\caption{Counts on the total number of url's returned by the search engine, and the total number of validated claims for Neymar, L. Messi and C. Ronaldo on the `diver' hypothesis, for both text and video entries. }
\label{tab:counts}
\vspace{-4pt}
\begin{center}
\begingroup\setlength{\fboxsep}{1pt}
\colorbox{gray!8}{%
\begin{tabular}{c | c | c | c | c | c | c } 
$\!\!$ENTITY$\!$ & $\!$MEDIA$\!$ & $\!$\#$\,$URL's$\!$ & $\!$\#$\,$CLAIMS$\!$ & $\!$YES$\!$ & $\!$NO$\!$ & $\!$\%YES$\!\!\!\!$\\
 \hline
Neymar & text & 579  & 71 & 55  & 16 & .77\\
Neymar & video & 891 & 63 &  62 & 01 & .98\\
\hline
L. Messi  & text & 546 & 128  & 32 & 96 & .25\\
L. Messi  & video & 643 & 63 &  21 & 42 & .33\\
\hline
$\!$C. Ronaldo$\!$  & text & 558 & 110  & 106 & 04 & .96\\
$\!$C. Ronaldo$\!$  & video & 849 & 103 &  95 & 08  & .92\\
\end{tabular}
}\endgroup
\end{center}
\vspace{-15pt}
\end{table}

\section{Experiment 2: Crowdsourcing Claims}\label{sec:crowd-claims}
\noindent
In this section we present the second experiment on our example hypothesis, `\emph{Does Neymar dive?}.' Now, instead of parsing claims mined from the web, we directly submitted videos to crowd workers and asked their rating w.r.t. the hypothesis.\footnote{Actually this second experiment has been conducted first, back in 2015.}  Our goal here is to get a different perspective on the problem, under more controlled conditions. 
Again, we will see if the data is suggestive that people may rate a hypothesis differently when given a particular piece of material evidence, than what they would say freely with no material at all.

\subsection{Experimental Setup}
\noindent
The experiment has been crowdsourced on July 18, 2015, at the \url{Microworkers.com} platform. For each task instance (rating of one video), we recruited 10 different workers. We evaluated a total of 11 videos (see next). By a greedy engagement of the crowd, in less than 2 hours all the 110 task instances were completed. 

\textbf{Crowd task design}. 
Fig.~\ref{fig:crowd-task} shows our design for this crowd task. Before seeing any video, the workers have been required to declare their prior bias on the hypothesis. A completed task was accepted only if the worker provided the correct answer for the `gotcha' sub-task shown at the bottom of the form about Neymar's T-shirts numbers. 

\textbf{Video sample selection}. We have searched for query `neymar dives' on a popular social media platform, and considered the videos highly ranked (top-30) by the search query. 
Striving for some opinion balance on the hypothesis, we have selected 11 videos, 
which are listed in Table \ref{tab:crowd} with both their social media and the crowdsourced statistics.

\begin{figure}[t]
\begin{subfigure}{.49\textwidth}
\begin{framed}
\vspace{-3pt}
\normalsize{\texttt{\textbf{$^\ast$}Soccer Hypothesis: Does Neymar Dive?\textbf{$^\ast$}}}\vspace{1pt}\\
\begin{scriptsize}
\noindent
\texttt{\vspace{-2.5pt}Soccer players "dive" if they frequently playact} \vspace{-2.5pt}$\!$\texttt{(cheat) unjustified falls on the pitch. So, does Neymar dive? What do you think?}\vspace{4pt}\\ 
\colorbox{cyan!45}{?} \hspace{3pt}\texttt{Whats is expected?}\vspace{2pt}\\ 
\texttt{1. Please inform your prior opinion: does Neymar dive?}\vspace{1pt}\\ 
\begin{tabular}{ll}
$\APLbox\!\!$ & \texttt{No}\\
$\APLbox\!\!$ & \texttt{Neutral}\\
$\APLbox\!\!$ & \texttt{Yes}\\
\end{tabular}\vspace{6pt}\\
\texttt{2. Please watch the video: \${MOVIELINK}.}\vspace{2pt}\\ 
\noindent
\texttt{3.\vspace{-2.5pt} Please give your vote based on $^\ast$that particular} \texttt{video$^\ast$: is Neymar diving?}
\vspace{5pt}\\
\begin{tabular}{ll}
$\APLbox\!\!$ & \texttt{No dives!}\\
$\APLbox\!\!$ & \texttt{Little bit.}\\
$\APLbox\!\!$ & \texttt{Sometimes.}\\
$\APLbox\!\!$ & \texttt{Many times.}\\
$\APLbox\!\!$ & \texttt{Massive dives!}\\
\end{tabular}\vspace{4pt}\\
\colorbox{red!50}{!} \hspace{3pt}\texttt{Required}:\vspace{0pt}\\
\texttt{1.$\!$ Your prior opinion. Example: No.}\vspace{-2pt}\\
\texttt{2.$\!$ 
Your vote based on the video. Example: Sometimes.} \vspace{-2pt}\\
\texttt{3.$\!$ \underline{Proof}: what numbers of Neymar's t-shirts appear in\vspace{-2.5pt} the video?}
\end{scriptsize}
\vspace{-3pt}
\end{framed}
\end{subfigure}
\caption{Crowd task design to collect opinions on the hypothesis about Neymar: declared prior bias and then a vote based on a specific video.}
\label{fig:crowd-task}
\vspace{-8pt}
\end{figure}

\begin{table*}[t]\footnotesize
\caption{Statistics of crowdsourced ratings on videos collected from a social media platform as material evidence for the hypothesis about Neymar. 
Each video (row) has been rated by 10 different workers. The two last columns show the average declared bias of these workers in contrast with their average rating for that specific video. Over all the 11 videos, we see (last row, .62 and .52) a shift of 10\% from the prior biases to the ratings on material evidence. }
\label{tab:crowd}
\vspace{-6pt}
\begin{center}
\begingroup\setlength{\fboxsep}{0.15pt}
\colorbox{gray!9}{%
\begin{tabular}{ | c | p{0.45\textwidth} | c | c || c | c | }
\hline
\multicolumn{4}{| c ||}{Social media data} & \multicolumn{2}{c |}{Crowdsourced data}\\
\hline
\hline
\multicolumn{2}{| c |}{$\!$user-generated material evidence$\!$} & $\!$http://www.youtube.com/watch?v=$\!$ & $\!$views$\!$ &$\,$avg. bias$\,$ & $\!$avg. rating$\!\!\!$\\
 \hline
 \rowcolor{white} $\!$1$\!$ & Neymar dive against South Korea? 8/8/2012 Olympic Games 2012 ... & \href{http://www.youtube.com/watch?v=qC3wmWTUDWw}{qC3wmWTUDWw} & 3.38K & .65 & .43\\
 \hline
$\!$2$\!$ & Neymar HORRIBLE Injury vs Colombia World Cup 04 07 2014 & \href{http://www.youtube.com/watch?v=BSMwAy_xbII}{BSMwAy\_xbII} & 61.54K & .55 & .43\\
 \hline
\rowcolor{white} $\!$3$\!$ & Neymar Crazy Dive - NO TOUCH!! Brazil vs Argentina ... 11/10/2014 & \href{http://www.youtube.com/watch?v=OZxkCJTTRps}{OZxkCJTTRps} & 0.49K & .60 & .40\\
 \hline
 $\!$4$\!$ & Neymar Jr HAS CHANGED $\cdot$ 2014-2015 ||HD|| & \href{http://www.youtube.com/watch?v=G0pPkSR4Fxw}{G0pPkSR4Fxw} & 0.33M & .70 & .48\\
 \hline
 \rowcolor{white} $\!$5$\!$ & Neymar Jr Horror Foul Bleeding Injury | Barcelona - Atletico Madrid & \href{http://www.youtube.com/watch?v=bEhVeFlP6w4}{bEhVeFlP6w4} & 0.11M & .60 & .48\\
 \hline
$\!$6$\!$ & Neymar - Horror Tackles \& Fights - Part 1 | HD & \href{http://www.youtube.com/watch?v=TiusCQv-3NM}{TiusCQv-3NM} & 0.26M & .50 & .50\\
 \hline
 \rowcolor{white}  $\!$7$\!$ & Neymar dive or no ?!! Up & \href{http://www.youtube.com/watch?v=pytMcZRJnKQ}{pytMcZRJnKQ} & 32.82K & .50 & .53\\
 \hline
 $\!$8$\!$ & Did Neymar Dive For Penalty & \href{http://www.youtube.com/watch?v=rpLMXXl4cV4}{rpLMXXl4cV4} & 0.56K & .75 & .60\\
 \hline
 \rowcolor{white} $\!$9$\!$ & How to stop Neymar Skills Fc Barcelona 2013 2014 ... & \href{http://www.youtube.com/watch?v=j3UyAR64evA}{j3UyAR64evA} & 1.98M
& .50 & .60\\
\hline
$\!$10$\!$ & Neymar Jr. - Goals\&Skills - 2011 - HD & \href{http://www.youtube.com/watch?v=IJGorSmEJM4}{IJGorSmEJM4} & 7.26M  & .55 & .50\\
 \hline
 \rowcolor{white} $\!$11$\!$ & Neymar $\cdot$ Best dives & \href{http://www.youtube.com/watch?v=UCM4b61nZpM}{UCM4b61nZpM} & 2.52M & .90 & .75\\
 \hline
 \hline
\multicolumn{3}{| c |}{TOTAL AVERAGE} & 1.14M & .62 & .52\\  
\hline
\end{tabular}
}\endgroup
\end{center}
\vspace{-6pt}
\end{table*}

\subsection{Crowdsourced Claims: Preliminary results}
\noindent
By looking at Table~\ref{tab:crowd} (see the bottom of the two last table columns), a comparison between the workers' average declared prior bias (normalized) and their average rating on each evidence piece shows a 10\% difference.

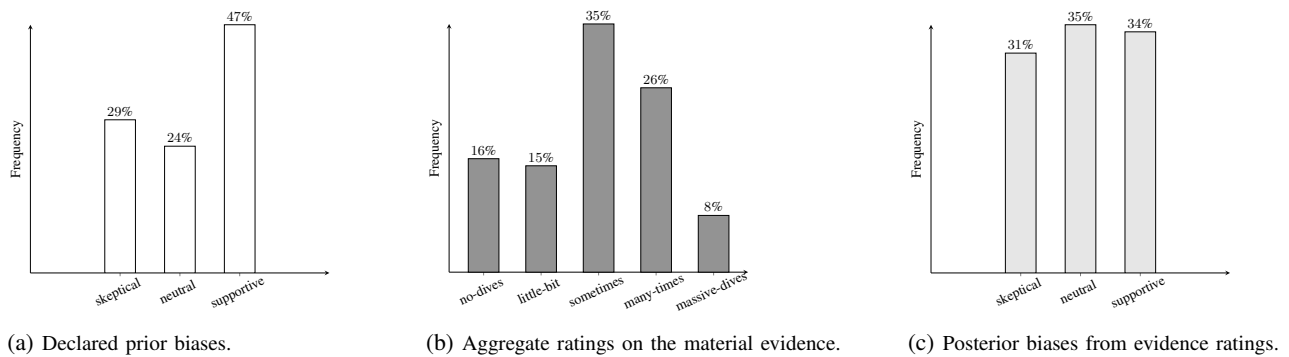
\begin{figure*}[t]\scriptsize
\advance\leftskip 0.8cm
\begin{subfigure}{0.3\textwidth}
\begin{tikzpicture}[font=\small,scale=0.58]
    \begin{axis}[
      ybar,
      bar width=20pt,
      ylabel={Frequency},
      ymin=0,
      ytick=\empty,
      xtick=data,
      axis x line=bottom, 
      axis y line=left,
      enlarge x limits=0.75,
      symbolic x coords={skeptical,neutral,supportive},
      xticklabel style={anchor=base,rotate=25,yshift=-0.40cm,xshift=-0.2cm},
      nodes near coords={\pgfmathprintnumber\pgfplotspointmeta\%}
    ]
      \addplot[fill=white, samples at={skeptical,neutral}] coordinates {
        (skeptical,29)
        (neutral,24)
        (supportive,47)
      };
    \end{axis}
\end{tikzpicture}
\caption{\footnotesize{Declared prior biases.}}\label{fig:bias}
\end{subfigure}
\hspace{5pt}
\begin{subfigure}{0.35\textwidth}
\begin{tikzpicture}[font=\small,scale=0.58]
    \begin{axis}[
      ybar,
      bar width=20pt,
      ylabel={Frequency},
      ymin=0,
      ytick=\empty,
      xtick=data,
      axis x line=bottom,
      axis y line=left,
      enlarge x limits=0.15,
      symbolic x coords={no-dives,little-bit,sometimes,many-times,massive-dives},
      xticklabel style={anchor=base,rotate=25,yshift=-0.40cm,xshift=-0.2cm},
      nodes near coords={\pgfmathprintnumber\pgfplotspointmeta\%}
    ]
      \addplot[fill=gray!85] coordinates {
        (no-dives,16)
        (little-bit,15)
        (sometimes,35)
        (many-times,26)
        (massive-dives,8)
      };
    \end{axis}
\end{tikzpicture}
\caption{\footnotesize{Aggregate ratings on the material evidence.}}
\label{fig:conf}
\end{subfigure}
\hspace{2pt}
\begin{subfigure}{0.32\textwidth}
\begin{tikzpicture}[font=\small,scale=0.58]
    \begin{axis}[
      ybar,
      bar width=20pt,
      ylabel={Frequency},
      ymin=0,
      ytick=\empty,
      xtick=data,
      axis x line=bottom, 
      axis y line=left,
      enlarge x limits=0.75,
      symbolic x coords={skeptical,neutral,supportive},
      xticklabel style={anchor=base,rotate=25,yshift=-0.40cm,xshift=-0.2cm},
      nodes near coords={\pgfmathprintnumber\pgfplotspointmeta\%}
    ]
      \addplot[fill=gray!20, samples at={skeptical,neutral}] coordinates {
        (skeptical,31)
        (neutral,35)
        (supportive,34)
      };
    \end{axis}
\end{tikzpicture}
\caption{\footnotesize{Posterior biases from evidence ratings.}}\label{fig:change}
\end{subfigure}
\caption{Histograms drawn from the crowdsourced data on the hypothesis about Neymar, for a total of 110 bias declarations and 110 video ratings.}
\label{fig:experiments}
\vspace{-8pt}
\end{figure*}

\textbf{Opinions shift on material evidence}. 
Fig.~\ref{fig:bias} shows an histogram of the workers' declared prior bias (skeptical, neutral, supportive). Fig. \ref{fig:conf} shows an histogram of their aggregate ratings on the evidence they were exposed to (no dives, little bit, sometimes, many dives, massive dives). 

The third histogram (Fig. \ref{fig:change}) is rendered by merging bins from the histogram of Fig. \ref{fig:conf}: the rates on bins `no-dives' and `little-bit' are turned into \emph{skeptical}, and the ones on bins `many-times' and `massive-dives' into \emph{supportive}. This is to convert the 1-5 scale of evidence rating onto a 1-3 scale to match the prior bias scale. Fig. \ref{fig:change} shows the result, the crowd's posterior average opinion based on evidence. 

If we compare the distributions of declared prior bias and of evidence rating, it does not look like an extraordinary change. Nevertheless, 
we see a shift from a more skewed prior bias (towards supporting the hypothesis) to a less skewed (more neutral) distribution after seeing evidence. This is suggestive that what people think may change after seeing a bunch of material evidence.

\section{Research Challenges}\label{sec:challenges}
\noindent
We discuss aspects that are not covered in this paper, but can be pursued in further research.

\textbf{Scientific method}. Hypotheses like `\emph{Neymar is a diver}' are `observational' \cite{losee2001}, as opposed to tentative explanations or complex theories. Their evaluation can only rely on the accumulated (material) evidence. Yet, given that evidence interpretation on controversial topics is intrinsically subjective and no groundtruth exists, the only available device for their assessment is still `people.' There are recent techniques (e.g., the Bayesian truth serum \cite{prelec2004}) to filter out people's biases, which can be useful in our framework.

\textbf{Statistical modeling}. Recall our assumption of a binary classification of claims. In fact, this is in view of modeling claims as Bernoulli trials.\footnote{See, e.g., [\url{http://en.wikipedia.org/wiki/Bernoulli_trial}].} We are pursuing rigorous statistical models for `hypothesis testing' on binomial counts.

\textbf{Hypothesis decision}. Given the evidence that has been accumulated in multiple forms (e.g., text and video), combining them into a weighted sum of scores towards a hypothesis decision is a core research problem. To establish an up-to-date decision, besides hypothesis testing methods, decision-theoretic frameworks may also be helpful \cite{robert2007}.

\textbf{Visual recognition}. The IBM Alchemy platform, e.g., has a component for `image tagging'.\footnote{\url{[http://www.alchemyapi.com/]}.} We have tried it on several images that are thumbnails for some of our extracted videos. This picture of Neymar (\href{http://www.oestadoonline.com.br/wp-content/uploads/2016/04/020116-Soccer-Barcelona-Neymar-PI-JE.vresize.1200.675.high_.80-300x169.jpg}{http://tinyurl.com/hyujhpe}), e.g., is tagged ``\emph{sport football soccer person futbol}". It does not tag `Neymar,' nor any property. Overall, our up-to-date report is that it cannot yet suit well, say, to the ideal task of automatic hypothesis evaluation based on material evidence. 

In fact, image classification may require training data. For instance, insurance companies can provide images to train classifiers of cars with and without a broken window (say, as positive and negative examples). Material evidence classification, however, is bound to be harder because it is intriguingly subjective and also tend to be a `new' problem for which labeled data is really unavailable --- both `subjectiveness' and `freshness' come by default, otherwise the target question may not qualify at all as a ``research'' problem. 
More investigation is needed to assess visual recognition techniques based on unsupervised learning.

\textbf{Speech recognition}. 
Audio checking, however, may be a more feasible research challenge. In this particular video data entry,\footnote{[http://www.youtube.com/watch?v=J13OPri3Ep0].} e.g., whose title is ``\emph{Neymar unfairly treated?},'' there is no explicit mention of the `diver' property in the snippet either. However, the speaker says ``\emph{Neymar has had a lot of criticism for his perceived diving. Now... Is he a diver or is he getting a rough treatment?}.'' So an interesting line of work is to extract the audio by speech recognition technology so that the claim gets amenable for NLP and could be validated. In this case, by solving the co-reference task that lies in it, viz., to match `he' in a sentence to `Neymar' in the other. 

A promising and cleaner strategy, when auxiliary media such as captions and subtitles are available, is to do NLP directly on text extracted from them. It may give better results than our approach in the first experiment (\S\ref{sec:web-claims}), since we extracted text from multimedia metadata (title and text snippet) that may not quite reflect the video content.

\section{Related Work}\label{sec:related-work}
\noindent
There is a fast-growing literature on detecting claims from text, usually on the web. However, we are not aware of prior work on the cognitive processing of multimedia data as material evidence for a hypothesis. Here is some related work on specific aspects that are worth discussing. 

\textbf{The Debator Project}. Our work is related to the so-called `debating' technologies at IBM \cite{rinott2015}. Their work is strongly based on machine learning methods to track correlated text passages that may be spread over a corpus. Going beyond the observational hypotheses that are more on our sight and are typical of social media, the goal is to enable eliciting all kind of textual information that may be relevant as evidence to decide on a claim. An example is: (topic) ``\emph{Use of performance enhancing drugs (PEDs) in professional sports}''; (claim) ``\emph{PEDs can be harmful to athletes health.}''; (evidence) ``\emph{The International Agency for Research on Cancer classifies androgenic steroids as ''Probably carcinogenic to humans.''}.'' To date, as far as we know this is yet a fully NLP project, which does not consider multimedia data for grounding on material evidence. 

\textbf{Sentiment analysis} \cite{liu2015}. The validation of natural language claims on arbitrary properties does not seem to suit well sentiment analysis techniques. We show why by means of an example from our collected web claims. The sentence ``\emph{Neymar Best Dive EVER!!!}'' receives positive sentiment, while the author clearly does not appreciate his perceived Neymar's behavior. So claim polarity detection taken as sentiment detection is misleading here. 

That is related with irony, which is a particular challenge for sentiment analysis that requires advanced techniques. In fact, the problem of detecting claims on a specific property for an entity is a somewhat different one, which is endowed with a pre-defined conceptual structure (see \S\ref{subsec:parsing}) and really seems to be better addressed directly by dependency parse trees. The preference of the latter over sentiment analysis techniques for the claim parsing and polarity detection task is discussed by Trummer et al. \cite{trummer2015}. 

\textbf{Question answering systems}. Propositional (yes/no) questions are an important class of questions that are addressed by Q\&A systems \cite{kanayama2012}, but our understanding of hypotheses as a specific kind of propositional questions is an important distinction in the enormous scope of Q\&A systems. Our focus is not on factoid-style questions, e.g., ``Was X born in year Y?.'' By `hypotheses' we mean the more subjective questions that are too intriguing to be freely answered, requiring some compelling (material) evidence.


\section{Conclusions}\label{sec:conclusions}

\noindent
In this paper we have presented two preliminary experiments as a ``baby'' step towards the evaluation of subjective hypotheses based on material evidence. If this line of research becomes feasible it will be an important application for question answering and cognitive computing in general. 

The key insight we have explored is that, although people may engage into never-ending discussions on such subjective topics, maybe on the interpretation of a particular piece of material evidence they can find themselves in agreement. At scale, if a data system accumulates evidence provided by several users, then cognitive computing could enable analytics and an up-to-date decision to answer such interesting subjective questions.  
We hope this becomes a new research front in multimedia cognitive computing. 

\section*{Acknowledgements}
\noindent
I owe the anonymous reviewers their feedback, which contributed to the final version of this paper. I am thankful to Rafael Brand\~ao and M\'arcio Moreno for their encouragement, and to Rog\'erio de Paula for recently pointing me to the ambitious Debator Project at IBM Resarch. I am indebted to Prof. H. V. Jagadish for reading and providing feedback on an earlier version of this manuscript in mid 2015.

\bibliographystyle{IEEEtran}
\bibliography{musdemp}

\begin{thebibliography}{10}
\providecommand{\url}[1]{#1}
\csname url@samestyle\endcsname
\providecommand{\newblock}{\relax}
\providecommand{\bibinfo}[2]{#2}
\providecommand{\BIBentrySTDinterwordspacing}{\spaceskip=0pt\relax}
\providecommand{\BIBentryALTinterwordstretchfactor}{4}
\providecommand{\BIBentryALTinterwordspacing}{\spaceskip=\fontdimen2\font plus
\BIBentryALTinterwordstretchfactor\fontdimen3\font minus
  \fontdimen4\font\relax}
\providecommand{\BIBforeignlanguage}[2]{{%
\expandafter\ifx\csname l@#1\endcsname\relax
\typeout{** WARNING: IEEEtran.bst: No hyphenation pattern has been}%
\typeout{** loaded for the language `#1'. Using the pattern for}%
\typeout{** the default language instead.}%
\else
\language=\csname l@#1\endcsname
\fi
#2}}
\providecommand{\BIBdecl}{\relax}
\BIBdecl

\bibitem{ferrucci2010}
D.~Ferrucci \emph{et~al.}, ``Building {W}atson: an overview of the {DeepQA}
  {P}roject,'' \emph{AI Magazine}, vol.~31, no.~3, 2010.

\bibitem{surowiecki2004}
J.~Surowiecki, \emph{The Wisdom of Crowds}.\hskip 1em plus 0.5em minus
  0.4em\relax Anchor, 2004.

\bibitem{trummer2015}
I.~Trummer, A.~Halevy, H.~Lee, S.~Sarawagi, and R.~Gupta, ``Mining subjective
  properties on the web,'' in \emph{ACM SIGMOD'15}, 2015, pp. 1745--60.

\bibitem{honnibal2015}
M.~Honnibal and M.~Johnson, ``An improved non-monotonic transition system for
  dependency parsing,'' in \emph{Proc. of the 2015 Conference on Empirical
  Methods in Natural Language Processing (EMNLP)}, 2015, pp. 1373--1378.

\bibitem{choi2015}
A.~S. J.~D.~Choi, J.~Tetreault, ``It depends: Dependency parser comparison
  using a web-based evaluation tool,'' in \emph{Proc. of the 53rd Annual
  Meeting of the ACL and the 7th International Joint Conf. on NLP}, 2015, pp.
  387--396.

\bibitem{choi2012}
J.~D. Choi and M.~Palmer, ``Guidelines for the {CLEAR} style constituent to
  dependency conversion guidelines for the clear style constituent to
  dependency conversion,'' University of Colorado, Tech. Rep. 01-12, 2012.

\bibitem{navarro2001}
G.~Navarro, ``A guided tour to approximate string matching,'' \emph{ACM
  Computing Surveys}, vol.~33, no.~1, pp. 31--88, 2001.

\bibitem{losee2001}
J.~Losee, \emph{A Historical Introduction to the Philosophy of Science},
  4th~ed.\hskip 1em plus 0.5em minus 0.4em\relax Oxford University Press, 2001.

\bibitem{prelec2004}
D.~Prelec, ``A bayesian truth serum for subjective data,'' \emph{Science}, vol.
  306, no. 5695, pp. 462--6, 2004.

\bibitem{robert2007}
C.~P. Robert, \emph{The Bayesian choice: From decision-theoretic foundations to
  computational implementation}, 2nd~ed., ser. Springer Texts in
  Statistics.\hskip 1em plus 0.5em minus 0.4em\relax Springer-Verlag, 2007.

\bibitem{rinott2015}
R.~Rinott, L.~Dankin, C.~Alzate, M.~M. Khapra, E.~Aharoni, and N.~Slonim,
  ``Show me your evidence -- an automatic method for context dependent evidence
  detection,'' in \emph{Proc. of the 2015 Conference on Empirical Methods in
  Natural Language Processing (EMNLP)}, 2015, pp. 440--450.

\bibitem{liu2015}
B.~Liu, \emph{Sentiment Analysis: Mining Opinions, Sentiments, and
  Emotions}.\hskip 1em plus 0.5em minus 0.4em\relax Cambridge University Press,
  2015.

\bibitem{kanayama2012}
H.~Kanayama, Y.~Miyao, and J.~Prager, ``Answering yes/no questions via question
  inversion,'' in \emph{Proc. of the International Conf. on Comp. Linguistics
  ({COLING})}, 2012, pp. 1377--92.

\end{thebibliography}

\end{document}